\documentclass[10pt, a4paper]{article}

\usepackage[final]{lrec2026} 

\usepackage{minted}
\usepackage{xcolor}
\usepackage{amsmath}
\usepackage{amssymb}
\usepackage{algorithmic}
\usepackage{algorithm}
\usepackage{bm}
\usepackage{booktabs}
\usepackage{comment}
\usepackage{multirow}
\usepackage{framed}
\usepackage{array}
\usepackage{arydshln}

\usepackage{cuted}
\usepackage{amsthm}
\usepackage{makecell}

\title{Multi-modal, Multi-task, Multi-criteria Automatic Evaluation\\
with Vision Language Models}

\name{Masanari Oi ${}^{1}$, Masahiro Kaneko ${}^{2, 1}$, Naoaki Okazaki ${}^{1, 3}$, Nakamasa Inoue ${}^{1}$} 

\address{${}^{1}$ Institute of Science Tokyo, ${}^{2}$ MBZUAI, ${}^{3}$ NII LLMC \\
         ohi.m.7b5f@m.isct.ac.jp}

\abstract{
Vision-language models (VLMs) have shown impressive abilities across a range of multi-modal tasks.
However, existing metrics for evaluating the quality of text generated by VLMs typically focus on an overall evaluation for a specific task, such as image captioning.
While the overall evaluation is essential for any task, the criteria prioritized can differ depending on the task, making it challenging for current metrics to adapt to multi-task scenarios.
To address this limitation, we propose HarmonicEval, a reference-free comprehensive evaluation metric that aggregates criterion-wise scores to produce the overall score in a bottom-up manner.
Furthermore, to assess the generalizability of automatic evaluation metrics in multi-task scenarios,  we construct the Multi-task Multi-criteria Human Evaluation (MMHE) benchmark, which comprises 18,000 expert human judgments across four multi-modal tasks.
Our experiments demonstrate that HarmonicEval achieves higher correlations with human judgments than conventional metrics while providing numerical scores for each criterion.
Project page: \url{https://stjohn2007.github.io/MMHE_project/}
 \\ \newline \Keywords{Automatic Evaluation, Annotated Dataset, Vision-Language models} }

\begin{document}

\maketitleabstract

\section{Introduction}
~\label{sec:intro}

\begin{figure}
\includegraphics[width=1.03\linewidth]{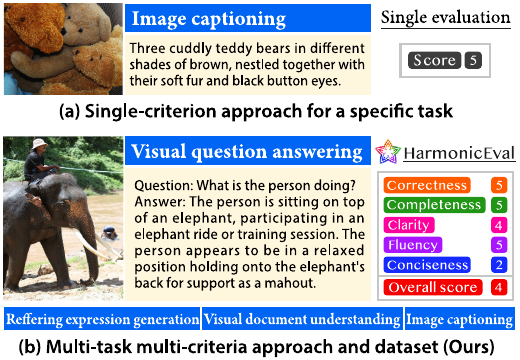}
\caption{
\textbf{Multi-task and multi-criteria evaluation.}
(a) Conventional single-criterion approach focuses on a single task, such as image captioning.
(b) HarmonicEval integrates multiple criteria to provide overall scores. MMHE consists of 18,000 expert human judgments across four multi-modal tasks and five criteria.
}
\label{fig:multi_criteria}
\end{figure}

Automatic evaluation of text generated by vision-language models (VLMs) is essential for improving their performance across various multi-modal tasks, such as image captioning and visual question answering~\cite{Hessel2021ClipS, lee-etal-2024-fleur}.
As the range of tasks that VLMs can perform continues to expand, developing specialized evaluation metrics for each task becomes increasingly difficult.
Hence, a comprehensive metric capable of evaluating text across multiple tasks is highly desirable.
However, most existing metrics focus on measuring the overall quality of text within a specific task, limiting their applicability in multi-task settings (see Figure~\ref{fig:multi_criteria}~(a)).

Existing metrics that provide only overall scores~\cite{Papineni2002bleu, Tianyi2020bertscore, Hessel2021ClipS} often prioritize specific evaluation criteria, as discussed in previous studies~\cite{kasai-etal-2022-transparent, fabbri-etal-2021-summeval}.
For example, metrics for evaluating image captions typically prioritize correctness and completeness over conciseness and fluency.
When these metrics are applied to other tasks, such as visual question answering, they tend to overvalue verbose or unnatural responses.
To address this limitation, integrating multiple evaluation criteria to predict the overall score, a concept we refer to as \textit{comprehensive evaluation}, holds significant potential for a more comprehensive assessment in multi-task scenarios.
However, this approach remains underexplored due to the lack of a meta-evaluation benchmark that provides human judgment 
across multiple tasks and criteria.
This motivates us to introduce a novel evaluation metric with a new benchmark in this paper.

\begin{figure*}[t]
\centering
\includegraphics[width=\linewidth]{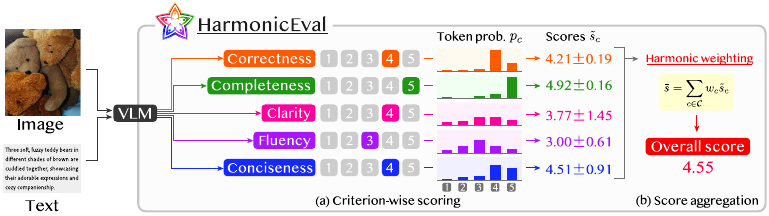}
\caption{HarmonicEval framework consists of two steps.
(a) Criterion-wise scoring is performed by prompting a VLM to evaluate the input text based on each criterion, followed by score smoothing to improve robustness based on the first-order statistics.
(b) Score aggregation produces an overall score using harmonic weighting based on the second-order statistics, aiming to reduce statistical fluctuations.
}
\label{fig:harmonic_eval}
\vspace{-3pt}
\end{figure*}

First, we propose HarmonicEval, a reference-free harmonic evaluation metric for multiple multi-modal tasks.
As shown in Figure~\ref{fig:multi_criteria}~(b), HarmonicEval integrates multiple criteria to produce the overall score in a bottom-up manner.
Specifically, the evaluation pipeline consists of two steps:
1) criterion-wise scoring, where a VLM is prompted to evaluate the input text based on each specific criterion,
and 2) score aggregation, where the overall score is calculated from the criterion-wise scores. For score aggregation, we introduce a novel harmonic weighting scheme that automatically determines weight coefficients based on the second-order statistics of the output token probability distributions.

Second, to assess the generalizability of automatic evaluation metrics, we introduce the multi-task multi-criteria human evaluation (MMHE) benchmark, the first meta-evaluation benchmark that provides human judgment annotations across multiple criteria and multiple multi-modal tasks.
Specifically, MMHE consists of 18,000 expert human judgments on five criteria for four diverse tasks: referring expression generation (REG), visual question answering (VQA), visual document understanding (VDU), and image captioning (IC).

Our experiments on MMHE show that HarmonicEval achieves higher correlations with human judgments than conventional metrics, while also providing criterion-specific scores that highlight areas for improvement. 
Furthermore, we demonstrate that HarmonicEval achieves state-of-the-art or comparable performance in conventional image caption evaluation scenarios across five widely used benchmarks that provide only overall judgments: Flickr8k-EX / CF~\cite{Hodosh2013Flickr8kExpert}, Composite~\cite{Aditya2018Composite}, PASCAL-50S~\cite{Vedantam2015cider}, and FOIL~\cite{shekhar-etal-2017-foil}.
In summary, our key contributions are threefold:
{
\vspace{-5pt}
\setlength{\leftmargini}{10pt}
\begin{itemize}
\setlength{\itemsep}{3pt}
\setlength{\parskip}{0pt}      
\item We propose HarmonicEval, a novel reference-free metric for harmonic evaluation across multiple multi-modal tasks.
\item We introduce MMHE, the first multi-task multi-criteria human evaluation benchmark, consisting of 18,000 expert judgments spanning four multi-modal tasks and five evaluation criteria.
\item We demonstrate the effectiveness of HarmonicEval on MMHE and five conventional image captioning benchmarks.
In addition, we conduct the first in-depth analysis of how existing metrics implicitly prioritize different evaluation criteria.
\end{itemize}
}

\section{HarmonicEval}
\label{sec:method}

As shown in Figure~\ref{fig:harmonic_eval}, the pipeline of HarmonicEval consists of two steps: criterion-wise scoring
(\hyperref[sec:method_prompting]{\textsection\ref*{sec:method_prompting}}) and score aggregation (\hyperref[sec:method_aggregation]{\textsection\ref*{sec:method_aggregation}}).

\subsection{Criterion-wise scoring}
\label{sec:method_prompting}

In this step, a VLM is employed as an evaluator and prompted to generate evaluation scores on each criterion independently.
Let $\bm{t}$ be an input text to be evaluated, such as an image caption for the IC task.
The evaluation process to obtain criterion-wise scores $s_{c}$ is formulated as $s_{c} = f([\bm{p}_{c}, \bm{t}], \bm{x})$,
where $c$ is a criterion, $\bm{x}$ is an input image, $f$ is a VLM, $\bm{p}_{c}$ is a prompt, and $[,]$ denotes textual concatenation.
To improve alignment with human judgments, score smoothing~\cite{liu2023geval,lee-etal-2024-fleur} is applied as $\tilde{s}_{c} = \sum_{r \in R} r P(r \mid [\bm{p}_{c}, \bm{t}], \bm{x})$ where $P(r \mid [\bm{p}_{c}, \bm{t}], \bm{x})$ is the output token probability of token $r$ assigned by the VLM, and $R$ is a set of ratings.

\subsection{Score aggregation}
\label{sec:method_aggregation}

To aggregate the criterion-wise scores $\tilde{s}_{c}$, we introduce harmonic weighting, a novel approach that leverages the second-order statistics of the output token probability distributions to adaptively determine the weight coefficients for aggregation.
Compared to simple averaging, our aggregation approach aims to better align with human evaluation by dynamically emphasizing more reliable scores based on the input.
Specifically, the overall score $S$ is computed as
\begin{align}
S
=
\sum_{c \in \mathcal{C}}
w_{c} \tilde{s}_{c},
\quad
w_{c} = \frac{1}{H}
\sigma_{c}^{- 2 (1-\gamma)/\gamma},
\end{align}
where $w_{c}$ is a weight coefficient, $\mathcal{C}$ is a set of criteria, 
$\sigma_{c}$ is the standard deviation of the criterion-wise score given by
\begin{align}
\sigma_{c} =
\sqrt{\sum_{r \in R} (r-\tilde{s}_{c})^{2}\ P(r \mid [\bm{p}_{c}, \bm{t}], \bm{x})},
\end{align}
and $H = \sum_{c \in \mathcal{C}} \sigma_{c}^{-2 (1-\gamma)/\gamma}$ is the harmonic mean of the variances with a hyperparameter $\gamma$.
Smaller values of $\sigma_{c}$ can be interpreted as indicating higher confidence in the evaluation of $c$.
The role of hyperparameter $\gamma$ is to bridge three weighting strategies: uniform weighting, inverse variance weighting and selective weighting as detailed below.

\paragraph{Uniform\hspace{1.5pt}weighting.}\hspace{1.5pt}When\hspace{1pt}$\gamma$\hspace{2pt}$=$\hspace{2pt}$1$, harmonic weighting reduces to uniform weighting $w_{c}$\hspace{2pt}$=$\hspace{2pt}$1/|\mathcal{C}|$. This is effective when all criterion-wise scores are equally reliable in determining the overall score.
However, this does not provide the best estimator as observed variances are ignored in aggregation.

\paragraph{Inverse\hspace{2pt}variance\hspace{2pt}weighting.} 
When $\gamma$\hspace{2pt}$=$\hspace{2pt}$0.5$, harmonic weighting reduces to inverse variance weighting $w_{c} \propto \sigma_{c}^{-2}$.
This provides the best linear unbiased estimator under the assumption that the observed variance is due to statistical fluctuations.
However, this assumption is not always reasonable, as each criterion may have its own variance.

\paragraph{Selective\hspace{2pt}weighting.} 
When $\gamma \to 0$, only the score $\tilde{s}_{c}$ with the smallest variance is selected as the overall score. This approach is used in experiments to show the necessity of aggregation.


\noindent \textbf{Discussion.} 
When the evaluation criteria are carefully designed, the uniform weighting ($\gamma = 1.0$) aligns closely with human expert judgment, and $0.5 \leq \gamma <1.0$ further improves the alignment because it adaptively reflects the confidence of criterion-wise scores.
Since the assumption underlying the inverse variance weighting ($\gamma=0.5$) is not reasonable
when each criterion has its own variance, we hypothesize that a value between 0.5 and 1.0 is optimal and choose $\gamma = 0.75$ as the default value.

\subsection{Implementation details}

\noindent \textbf{Definition of criteria.} 
Based on prior research in natural language generation~\cite{asano-etal-2017-reference, kryscinski-etal-2019-neural, fabbri-etal-2021-summeval, freitag-etal-2021-experts, song-etal-2024-finesure} and multi-modal evaluation~\cite{Aditya2018Composite, kasai-etal-2022-transparent}, we define five evaluation criteria: $\mathcal{C}\hspace{-3pt}=\hspace{-3pt}\{\text{Correctness},$ $\text{Completeness}, \text{Fluency},$\hspace{1pt}$\text{Conciseness}, \text{Clarity}\hspace{0pt}\}$.
Their definitions are summarized in Table~\ref{tab:criteria}.
We validated these criteria by examining 100 outputs from various vision-language tasks and confirmed their adequacy for reliable evaluation.

Aggregating these five criteria contributes to the overall text quality evaluation across a wide range of tasks. Depending on the task and the style of the input text, some criteria may not be necessary.
However, when such evaluations are conducted, the output becomes less confident, leading to higher variance $\sigma_{c}$ and lower weight coefficients $w_{c}$, as VLMs account for task and criterion features in addition to the input text.
Thus, HarmonicEval can adaptively perform comprehensive evaluations without manual tuning of weight coefficients.

\begin{table}[t]
\centering
\small
\begin{tabular}{p{0.95\columnwidth}}
\toprule
\textbf{Correctness (Crt):} The degree to which the target text accurately reflects the content of the input image and text.\\ 
\midrule
\textbf{Completeness (Cmp):} The extent to which the target text captures all relevant and significant details of the input image and text. \\
\midrule
\textbf{Clarity (Clr):} The ease with which the reader can understand the target text.\\ 
\midrule
\textbf{Fluency (Flu):} The grammatical accuracy and natural flow of the target text.\\ 
\midrule
\textbf{Conciseness (Cnc):} The efficiency of the target text in conveying information without unnecessary verbosity.\\ 
\bottomrule
\end{tabular}
\caption{Five criteria for HarmonicEval and MMHE.\label{tab:criteria}}
\vspace{-3pt}
\end{table}

\begin{figure*}[t]
\centering
\includegraphics[width=0.98\linewidth]{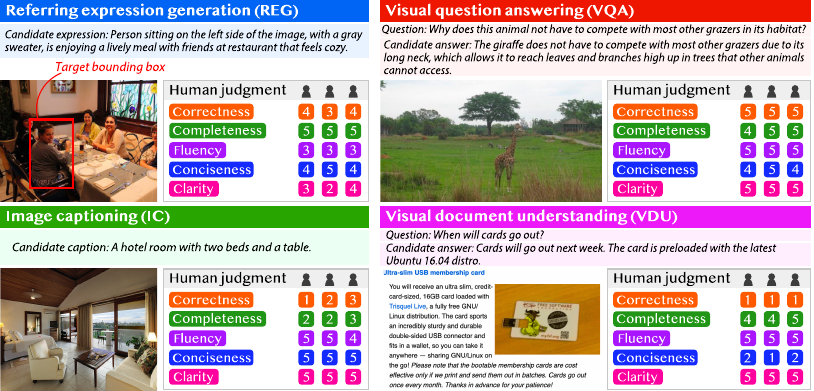}
\caption{MMHE benchmark is a multi-task multi-criteria human evaluation benchmark. Each candidate text is manually evaluated by three expert annotators.}
\label{fig:mmhe_examples}
\end{figure*}

\noindent \textbf{Prompts.} 
The prompt $\bm{p}_{c}$ instructs the VLM to evaluate text with respect to the criterion $c$ on a five-point scale $R = \{1,2,3,4,5\}$.
Below is the prompt of the correctness criterion for the IC task.

\noindent\textit{Your task is to rate the \textbf{caption} for the given image on a scale of 1 to 5 on the following criterion and rating scale.}

\noindent\textit{Evaluation Criterion: Correctness}

\noindent\textit{How accurately does \textbf{the caption describe the image?}}

\noindent\textit{Rating Scale:}

\noindent\textit{- 1 Very Low Correctness: The \textbf{caption} is mostly or entirely incorrect ...}

\noindent\textit{...}

\noindent\textit{- 5 Extremely High Correctness: The \textbf{caption} perfectly captures all ...}

Here, the boldfaced portions indicate the task-dependent phrases. For example, ``caption'' is replaced with ``answer'' for the VQA task.

\section{MMHE Benchmark}
\label{sec:dataset}
We present the MMHE benchmark, the first meta-evaluation benchmark that covers multiple evaluation criteria across multiple multi-modal tasks.
We collected 18,000 expert human judgments spanning four tasks and five criteria.
Example human judgment scores are shown in Figure~\ref{fig:mmhe_examples}.

\subsection{Benchmark design}
\label{sec:dataset_design}

\paragraph{Motivation.}
Despite numerous human evaluation benchmarks for multi-modal tasks, most focus on a single task (e.g., image captioning) or offer only an overall quality label. 
Our purpose is twofold: 
1) to assess how evaluation metrics perform in a multi-task setting, and 
2) to investigate how metrics that provide only overall scores prioritize certain criteria over others.
To achieve these goals, we design MMHE to cover multiple multi-modal tasks and to explicitly collect human judgments across five evaluation criteria.

\paragraph{Multi-modal tasks.}
We select four diverse tasks to show how different criteria matter across contexts:
1) \textbf{REG} aims to generate a textual expression that uniquely identifies a specific object in the image (marked by a bounding box). We expect completeness to be crucial for precisely distinguishing the target object.
2) \textbf{VQA} requires generating an answer to a question about the image content. Given the nature of question answering, we hypothesize that correctness and conciseness are particularly important here.
3) \textbf{VDU} focuses on interpreting information from visually presented documents. Similar to VQA, we suspect that correctness and conciseness play key roles.
4) \textbf{IC} involves producing a descriptive sentence for the entire image. We anticipate that correctness and completeness are especially relevant for capturing key elements.

MMHE is the first benchmark to integrate multiple multi-modal tasks with a unified set of evaluation criteria in a single framework.
This design enables fine-grained, criterion-wise analysis of how different metrics perform across various task requirements, which cannot be achieved by simply combining existing benchmarks with their disparate evaluation criteria.

\subsection{Benchmark construction}\label{sec:data_construction}
The benchmark construction process consists of three steps: 1) Source selection, which selects source text-image pairs;
2) Target generation, which creates target texts to be evaluated using state-of-the-art VLMs; and
3) Human expert evaluation, which assesses the quality of the target texts.

\paragraph{Source selection.}
We selected the following four datasets: RefCOCO \cite{kazemzadeh-etal-2014-referitgame} for REG, OK-VQA \cite{Marino2019OKVQAAV} for VQA, VisualMRC \cite{VisualMRC2021} for VDU, and MSCOCO \cite{Lin2014MicrosoftCC} for IC.
We randomly sampled 100 instances from the validation or test subset of each dataset.

\paragraph{Target generation.}
The target texts to be evaluated were generated using state-of-the-art VLMs.
Specifically, we employed ten VLMs: LLaVA-1.5-7B/13B~\cite{liu2023llava}, InstructBLIP-Vicuna-7B/13B~\cite{Dai23InstructBLIP}, Qwen-VL~\cite{bai2023qwenvl}, Qwen2-VL-Instruct-7B/72B~\cite{Qwen2VL}, CogVLM-Chat~\cite{wang2024cogvlm}, GPT-4o-mini and GPT-4o~\cite{openai2024gpt4ocard}.
For each instance, we assigned three distinct VLMs from this pool, ensuring that every instance had exactly three candidate outputs. The assignment was balanced across tasks and models.
This design resulted in 100 instances × 3 outputs = 300 candidate responses per task.

\paragraph{Human expert evaluation.}
Five expert annotators were given an explanation of the four multi-modal tasks and asked to carefully review the five evaluation criteria in Table~\ref{tab:criteria}.
They rated each target text on a five-point scale, then conducted a thorough review for consistency.
Each instance was independently scored by three annotators, yielding 18,000 human judgment scores.

\paragraph{Overall judgments.}
To collect overall judgments, we adopted a best-of-three approach, in which annotators choose the best output among three responses generated by different models for the same input. We did not use a five-point scale for the overall score because defining each rating level (from 1 to 5) for the overall quality is difficult and could introduce bias~\cite{chiang2024chatbotarenaopenplatform}.

\begin{figure}
\centering
\includegraphics[width=1.0\columnwidth]{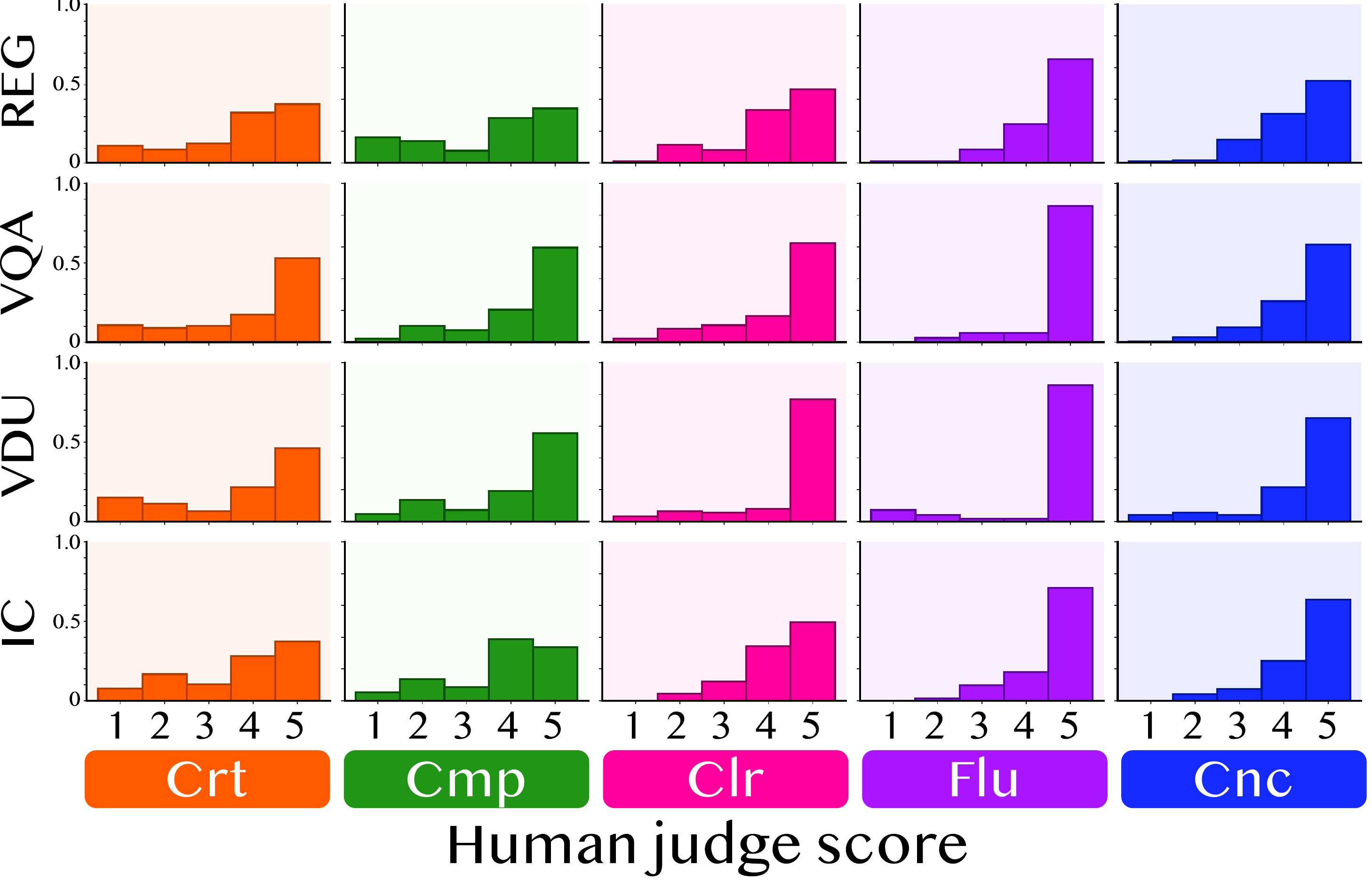}
\caption{Human judgment score distributions for each task and criterion on the MMHE benchmark.}
\label{fig:statistics}
\end{figure}

\subsection{Data analysis}
\label{sec:data_analysis}
Figure~\ref{fig:statistics} presents the score distributions for each criterion across the four tasks.
The correctness and completeness criteria exhibit diverse score distributions for most tasks, suggesting that even the state-of-the-art VLMs face challenges in these aspects. 
In contrast, the fluency criterion shows a narrow score distribution across most tasks, with a dominant score of five. 
This indicates that the VLMs generate fluent text even when visual understanding is inaccurate.
Nonetheless, we consider fluency an essential criterion, as overlooking it could lead to high scores being assigned to correct but non-fluent text.
A qualitative example illustrating this is provided in Section~\ref{sec:analysis}.
The clarity and conciseness criteria have score distributions that are intermediate between the diverse and narrow distributions.
While the score of five is more prevalent than in the diverse distributions, other scores still occur with noticeable frequency.

\begin{table}[t]
\centering
\small
\tabcolsep 6pt
\begin{tabular}{lccccc}
\toprule
Method & REG & VQA & VDU & IC & Avg. \\
\midrule
BLEU & $45.3$ & $29.4$ & $57.3$ & $46.8$ & $44.7$ \\
ROUGE & $49.0$ & $30.8$ & $56.0$ & $47.9$ & $45.9$ \\
CIDEr & $42.5$ & $25.0$ & $62.1$ & $42.7$ & $43.1$ \\
METEOR & $44.4$ & $29.4$ & $59.7$ & $53.6$ & $46.8$ \\
\midrule
BERT-S & $46.2$ & $33.8$ & $62.1$ & $53.1$ & $48.8$ \\
BART-S & $56.4$ & $20.5$ & $60.9$ & $57.8$ & $48.9$ \\
CLIP-S & $60.1$ & $39.7$ & $60.9$ & $52.0$ & $53.2$ \\
G-VEval & $60.1$ & $75.0$ & $71.9$ & $68.7$ & $68.9$ \\
FLEUR & $62.9$ & $\bm{76.4}$ & $60.9$ & $73.9$ & $68.5$ \\
GPT-FLEUR & $60.1$ & $75.0$ & $\bm{76.5}$ & $76.0$ & $71.9$ \\
HarmonicEval & $\bm{66.6}$ & $\bm{76.4}$ & $73.4$ & $\bm{77.0}$ & $\bm{73.4}$ \\
\bottomrule
\end{tabular}
\caption{
Accuracy (\%) on MMHE.
The best result for each task is marked in bold. Average (Avg.) indicates the average accuracy across the four tasks.
}
\label{tab:sota}
\end{table}

\def\bmm#1{#1}
\def\highscore#1{\textcolor{red}{\underline{#1}}}
\def\lowscore#1{\textcolor{blue}{\underline{#1}}}
\def\minus{$-$}

\begin{table*}[t]
\centering
\small
\tabcolsep 1.3pt
\begin{tabular}{l!{\vrule}ccccc!{\vrule}ccccc!{\vrule}ccccc!{\vrule}cccccc}
\toprule
& \multicolumn{5}{c!{\vrule}}{REG} & \multicolumn{5}{c!{\vrule}}{VQA} & \multicolumn{5}{c!{\vrule}}{VDU} & \multicolumn{5}{c}{IC} \\
Metric & Crt & Cmp & Clr & Flu & Cnc & Crt & Cmp & Clr & Flu & Cnc & Crt & Cmp & Clr & Flu & Cnc & Crt & Cmp & Clr & Flu & Cnc \\
\midrule
BLEU & $6.0$ & \highscore{$6.9$} & $3.9$ & \lowscore{$1.2$} & $6.1$ & $\minus1.3$ & $\minus10.4$ & $\minus11.0$ & \lowscore{$\minus19.3$} & \highscore{$4.1$} & $19.8$ & \lowscore{$12.9$} & $14.9$ & $14.3$ & \highscore{$21.2$} & $4.4$ & $4.5$ & $5.9$ & \lowscore{$0.3$} & \highscore{$11.3$} \\
ROUGE & $2.3$ & \highscore{$5.7$} & $4.4$ & \lowscore{$\minus3.5$} & $3.9$ & $7.1$ & $\minus2.8$ & $\minus5.0$ & \lowscore{$\minus8.1$} & \highscore{$10.2$} & $20.0$ & \lowscore{$14.7$} & $16.2$ & $17.9$ & \highscore{$22.7$} & $5.2$ & $6.5$ & $9.0$ & \lowscore{$4.4$} & \highscore{$9.7$} \\
CIDEr & $6.4$ & $3.4$ & $2.4$ & \lowscore{$\minus9.7$} & \highscore{$20.9$} & $\minus27.8$ & \lowscore{$\minus39.0$} & $\minus19.5$ & $\minus26.0$ & \highscore{$\minus3.8$} & $23.7$ & \lowscore{$15.8$} & $19.3$ & $18.0$ & \highscore{$23.8$} & $0.7$ & $\minus1.6$ & $8.7$ & \lowscore{$\minus3.8$} & \highscore{$14.5$} \\
METEOR & $1.9$ & \highscore{$5.3$} & $5.2$ & $\minus5.1$ & \lowscore{$\minus6.3$} & \highscore{$5.3$} & $\minus3.9$ & $\minus8.2$ & \lowscore{$\minus8.5$} & $2.7$ & $17.8$ & $18.0$ & $16.9$ & \highscore{$20.5$} & \lowscore{$14.9$} & $6.8$ & \highscore{$12.1$} & $7.3$ & \lowscore{$\minus2.3$} & $1.0$ \\
\midrule
BERT-S & $6.5$ & $6.9$ & $\minus6.5$ & \lowscore{$\minus8.6$} & \highscore{$12.4$} & $\minus2.8$ & \lowscore{$\minus14.3$} & $4.9$ & $\minus10.0$ & \highscore{$6.1$} & $21.0$ & \lowscore{$17.4$} & $20.4$ & $21.6$ & \highscore{$23.9$} & \highscore{$12.3$} & $11.1$ & $6.4$ & \lowscore{$4.7$} & $10.5$ \\
BART-S & $4.4$ & \highscore{$6.7$} & $4.2$ & \lowscore{$\minus7.8$} & $3.1$ & $\minus13.4$ & \lowscore{$\minus20.2$} & $\minus2.8$ & $\minus16.6$ & \highscore{$1.6$} & \highscore{$22.4$} & $21.3$ & $21.6$ & $17.9$ & \lowscore{$14.7$} & \highscore{$4.8$} & $4.3$ & $4.3$ & \lowscore{$2.2$} & $3.2$ \\
CLIP-S & $13.5$ & \highscore{$14.4$} & $6.8$ & $\minus0.9$ & \lowscore{$\minus5.1$} & $6.6$ & $5.4$ & $7.2$ & \highscore{$8.1$} & \lowscore{$4.5$} & \highscore{$15.2$} & $12.5$ & $15.0$ & $12.6$ & \lowscore{$8.4$} & $20.2$ & \highscore{$21.3$} & $11.1$ & \lowscore{$3.2$} & $3.5$ \\
G-VEval & $11.4$ & \highscore{$23.4$} & $18.7$ & $9.9$ & \lowscore{$8.3$} & \highscore{$52.8$} & $41.0$ & \highscore{$19.0$} & $44.7$ & $35.9$ & \highscore{$54.1$} & $41.7$ & $47.0$ & \lowscore{$40.6$} & $42.0$ & \highscore{$43.8$} & $43.4$ & $21.9$ & $26.0$ & \lowscore{$14.5$} \\
FLEUR & $\bm{29.3}$ & \highscore{$\bm{30.8}$} & $18.6$ & \lowscore{$8.7$} & $11.2$ & $38.7$ & \lowscore{$38.2$} & $\bm{39.9}$ & $39.8$ & \highscore{$\bm{44.7}$} & $38.1$ & $37.1$ & \highscore{$44.6$} & $35.2$ & \lowscore{$28.2$} & $33.9$ & \highscore{$35.0$} & $\bm{25.9}$ & $24.5$ & \lowscore{$14.0$} \\
GPT-FLEUR & $19.7$ & \highscore{$30.6$} & $14.0$ & $19.7$ & \lowscore{$11.0$} & \highscore{$\bm{54.7}$} & $42.0$ & \lowscore{$18.0$} & $35.0$ & $23.0$ & \highscore{$59.0$} & $44.4$ & $43.8$ & $37.2$ & \lowscore{$29.5$} & $\bm{47.5}$ & \highscore{$47.7$} & $25.3$ & $29.1$ & \lowscore{$13.4$} \\
\midrule
HarmonicEval & $23.2$ & $\bm{30.8}$ & $\bm{24.0}$ & $\bm{20.7}$ & $\bm{23.8}$ & $53.5$ & $\bm{50.6}$ & $31.8$ & $\bm{51.9}$ & $44.4$ & $\bm{60.0}$ & $\bm{48.8}$ & $\bm{47.9}$ & $\bm{51.2}$ & $\bm{45.8}$ & $44.7$ & $\bm{50.3}$ & $19.8$ & $\bm{36.4}$ & $\bm{22.8}$ \\
\bottomrule
\end{tabular}
\caption{
Criterion-wise correlation analysis on MMHE.
Scores for the most positively and negatively correlated criteria are marked with red and blue underlines, respectively.
The highest correlations for each criterion are highlighted in bold.
Crt: Correctness,
Cmp: Completeness,
Clr: Clarity,
Flu: Fluency,
Cnc: Conciseness.}
\label{tab:correlation_analysis}
\end{table*}

\section{Experiments}

\subsection{Performance on MMHE}

We evaluate the performance of HarmonicEval on MMHE and compare it with conventional metrics.
We also examine how existing metrics prioritize or deprioritize specific criteria in each task.

\paragraph{Settings.}
To assess the performance of each metric, we use accuracy (\%) for the overall evaluation and the Kendall's tau correlation coefficient $\tau$ for criterion-wise evaluations.

We implement nine baselines, including four n-gram-based metrics~\cite{Papineni2002bleu, Lin2004rouge, Vedantam2015cider, Banerjee2005METEOR} and five neural network-based metrics~\cite{Tianyi2020bertscore, NEURIPS2021_e4d2b6e6, Hessel2021ClipS, Tong_He_Shao_Yeung_2025, lee-etal-2024-fleur}, grouped in Table~\ref{tab:sota}.
Among them, FLEUR~\cite{lee-etal-2024-fleur} is a state-of-the-art VLM-based metric\footnote{Note that all baseline metrics don't support criterion-wise scoring and produce only overall scores.}.
HarmonicEval utilizes GPT-4o as its backbone.
For a fair comparison with FLEUR, we also implement GPT-FLEUR, which substitutes GPT-4o for the original LLaVA-1.5-13B.

\paragraph{Main results.}
Table~\ref{tab:sota} compares HarmonicEval with conventional metrics in terms of overall performance on MMHE.
HarmonicEval achieves the highest accuracy in REG (66.6), VQA (76.4), IC (77.0), and attains the top average score of 73.4 across tasks.
While GPT-FLEUR obtains the highest score on VDU (76.5), it performs less effectively on REG.
These results underscore the strong multi-task capability of HarmonicEval.

\paragraph{Correlation analysis.}
To investigate how existing metrics prioritize or deprioritize certain criteria, we show the correlation between the predicted overall scores and the human judgment scores for each criterion in Table~\ref{tab:correlation_analysis}.
In the table, red and blue underlines denote the most and least correlated criterion, respectively, for each task.

We observe task-wise trends. For REG, completeness shows the highest correlation across most metrics. This is reasonable, as REG requires explicit expressions to identify a unique object by distinguishing it from marked objects.
For VQA, most metrics are more strongly correlated with conciseness but less so with completeness.
This indicates that conventional metrics deprioritize completeness, potentially leading to inaccurate evaluations of insufficient answers.  
A similar trend is observed in VDU, where completeness is also deprioritized by conventional metrics.
For IC, fluency exhibits low correlations for most metrics, suggesting a tendency to assign high scores even to non-fluent texts.
Overall, these results underscore the necessity of a comprehensive evaluation metric in multi-task scenarios.

Table~\ref{tab:correlation_analysis} also shows the correlations between the criterion-wise scores of HarmonicEval and the human judgment scores.
HarmonicEval achieves the highest correlation across most criteria.
While this is expected, as HarmonicEval is the only metric that outputs criterion-wise scores, the result nevertheless demonstrates that its predictions align well with human judgments on each individual criterion.

\subsection{Analysis}
\label{sec:analysis}

\begin{figure*}[t]
    \centering
    \includegraphics[width=\linewidth]{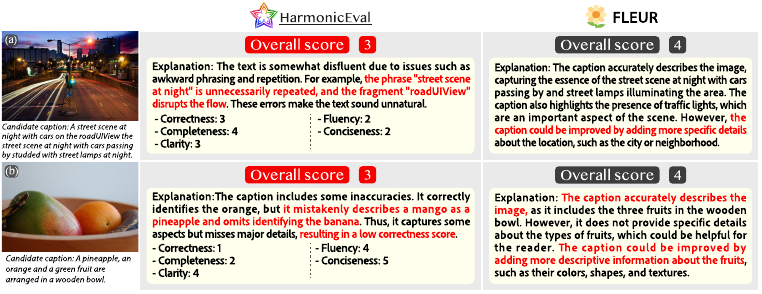}
    \caption{Qualitative examples.}
    \label{fig:qualitative}
\end{figure*}

\paragraph{Can HarmonicEval improve explainability?}

Providing feedback on evaluation results to users is important.
To investigate whether HarmonicEval offers clear textual explanations, we prompt the VLM with ``\textit{Why? Tell me the reason.}'' after obtaining the overall scores, following \citet{lee-etal-2024-fleur}.

Figure~\ref{fig:qualitative} shows a qualitative comparison of HarmonicEval and FLEUR on the IC task.
Panel (a) shows that HarmonicEval successfully detects severe fluency issues and reflects them in its overall score, whereas FLEUR tends to overlook such deficiencies.
This suggests that FLEUR may fail to capture certain criteria (e.g., fluency), as discussed in Section~\ref{sec:data_analysis}.
Panel (b) illustrates that HarmonicEval identifies incorrect details in captions more precisely than FLEUR, assigning a lower correctness score accompanied by a criterion-specific explanation.
These examples indicate that HarmonicEval provides more informative textual feedback.

To verify this observation more systematically, we conduct a user study to quantitatively assess the explainability of HarmonicEval.
Specifically, we sample 25 instances from each task and generate explanations for both HarmonicEval and FLEUR, yielding 100 explanation pairs in total.
Five human annotators then evaluate which explanation in each pair is more informative and useful, allowing for ties.
As shown in Table~\ref{tab:userstudy}, HarmonicEval significantly outperforms FLEUR on textual explainability, consistent with its more fine-grained criterion-wise evaluation observed in the qualitative examples.

\begin{table}[t]
\centering
\small
        \begin{tabular}{cccccc}
        \toprule
         & REG & VQA & VDU & IC & Total \\
        \midrule
        HarmonicEval  & \bm{$19^{*}$} & \bm{$12$} & \bm{$21^{*}$} & \bm{$19^{*}$} & \bm{$71^{*}$} \\
        FLEUR  & $3$ & $9$ & $3$ & $3$ & $18$ \\
        Tie & $3$ & $4$ & $1$ & $3$ & $11$ \\
        \bottomrule
        \end{tabular}
        \caption{User study on textual explainability. Asterisks ($*$) denote statistical significance between HarmonicEval and FLEUR ($p < 0.05$, binomial test).}
        \label{tab:userstudy}
\end{table}

\begin{table}[t]
\centering
\small
\tabcolsep 2pt
\begin{tabular}{lccccc}
        \toprule
        Metric & REG & VQA & VDU & IC & Avg. \\
        \midrule
        HarmonicEval  & $\bm{66.6}$ & $\bm{76.4}$ & $73.4$ & $\bm{77.0}$ & $\bm{73.4}$ \\
        \scalebox{0.97}[1]{w/o criterion-wise scoring} & $62.0$ & $73.5$ & $\bm{75.9}$ & $76.5$ & $72.0$ \\
        \scalebox{0.97}[1]{w/o harmonic weighting} & $65.7$ & $75.0$ & $73.4$ & $76.5$ & $72.6$ \\
        \bottomrule
        \end{tabular}
        \caption{Ablation study.
        }
        \label{tab:ablation}
\end{table}

\begin{table}[t]
\centering
\small
        \begin{tabular}{cccccc}
        \toprule
        $\gamma$ & REG & VQA & VDU & IC & Avg. \\
        \midrule
        $0.01$  & $47.2$ & $69.1$ & $61.4$ & $53.1$ & $57.7$ \\
        $0.50$  & $\bm{66.6}$ & $\bm{76.4}$ & $\bm{73.4}$ & $76.5$ & $73.2$ \\
        $0.75$  & $\bm{66.6}$ & $\bm{76.4}$ & $\bm{73.4}$ & $\bm{77.0}$ & $\bm{73.4}$ \\
        $1.00$  & $65.7$ & $75.0$ & $\bm{73.4}$ & $76.5$ & $72.6$ \\
        \bottomrule
        \end{tabular}
        \caption{\scalebox{0.97}[1]{Hyperparameter study.}}
        \label{tab:gamma_results}
\end{table}

\begin{table}[t]
    \centering
    \small
    \setlength{\tabcolsep}{3pt}
    \begin{tabular}{llccccc}
    \toprule
    VLM & Method & REG & VQA & VDU & IC & Avg. \\
    \midrule
    \multirow{2}{*}{L-7B}
    
    & FLEUR & $\bm{69.4}$ & $69.1$ & $\bm{63.2}$ & $72.6$ & $\bm{68.6}$ \\
    & Harmonic & $62.9$ & $\bm{72.0}$ & $61.4$ & $\bm{73.4}$ & $67.4$ \\
    \midrule
    \multirow{2}{*}{L-13B} 
    & FLEUR & $62.9$ & $76.4$ & $60.9$ & $\bm{73.9}$ & $68.5$ \\
    & Harmonic & $\bm{64.8}$ & $\bm{77.9}$ & $\bm{63.2}$ & $72.9$ & $\bm{69.7}$ \\
    \midrule
    \multirow{2}{*}{G-4o} 
    & FLEUR & $60.1$ & $75.0$ & $\bm{76.5}$ & $76.0$ & $71.9$ \\
    & Harmonic & $\bm{66.6}$ & $\bm{76.4}$ & $73.4$ & $\bm{77.0}$ & $\bm{73.4}$ \\
    \bottomrule
    \end{tabular}
    \caption{Comparison of HarmonicEval and FLEUR across different backbone VLMs.}
    \label{tab:llm_acc}
\end{table}

\paragraph{Is each component in HarmonicEval essential?}
We conduct an ablation study on the two key components of HarmonicEval: criterion-wise scoring and harmonic weighting.
Specifically, we examine two alternative approaches:
1) prompting the VLM to directly predict an overall score based on all five criteria, without using criterion-wise scores (\textit{w/o criterion-wise scoring}); and 2) computing the overall score as a simple average of the criterion-wise scores (\textit{w/o harmonic weighting}).

Table~\ref{tab:ablation} shows that both components improve overall performance. 
Removing criterion-wise scoring lowers scores on REG, VQA, and IC, indicating that explicitly scoring each criterion results in better evaluation than relying on a single overall score based on a detailed prompt.
Similarly, omitting harmonic weighting reduces performance in most tasks, validating the effectiveness of our approach.

\paragraph{Is statistical aggregation effective?}
Table~\ref{tab:gamma_results} shows a hyperparameter study for $\gamma$. As expected, harmonic weighting with $\gamma = 0.75$ performs the best.
This justifies the importance of the statistical aggregation process.


\paragraph{Is HarmonicEval effective across various backbone VLMs?}

Table~\ref{tab:llm_acc} shows the overall performance comparison between HarmonicEval and FLEUR using LLaVA-1.5-7B, LLaVA-1.5-13B, and GPT-4o as backbone models. 
HarmonicEval consistently outperforms FLEUR on both LLaVA-1.5-13B and GPT-4o.
On the other hand, FLEUR slightly outperforms HarmonicEval when using LLaVA-1.5-7B.
We found that this is because LLaVA-1.5-7B tends to underrate texts in the conciseness criterion compared to human-assigned scores.
Nonetheless, these results highlight HarmonicEval's effectiveness across different VLMs, particularly with more capable ones.

\subsection{Performance on existing IC benchmarks} \label{sec:existing}
To assess the robustness of HarmonicEval in standard image captioning scenarios, we also evaluate it on five widely used IC benchmarks: Flickr8k-EX / CF, Composite, Pascal-50S, and FOIL.
They support comparisons with a broader set of baselines, including specialized IC metrics~\cite{Anderson2016spice, Jiang2019tiger, lee-etal-2020-vilbertscore, hu2023infometic, Wada2024PolosPolaris}.
Following prior works~\cite{Hessel2021ClipS, lee-etal-2024-fleur}, we use Kendall’s tau-c for Flickr8k-EX and Composite, tau-b for Flickr8k-CF, and accuracy for Pascal-50S.

As shown in Tables~\ref{tab:score} and \ref{tab:foil}, HarmonicEval matches or surpasses state-of-the-art metrics on Flickr8k-CF, Composite, and FOIL.
Despite modest gaps on Flickr8k-EX and Pascal-50S relative to computationally heavier or fine-tuned, task-specific metrics, strong results on the remaining three benchmarks underscore HarmonicEval’s overall robustness without task-specific tuning.

\begin{table}
\tabcolsep 5pt
\small
\centering
\begin{tabular}{llcccccc}
\toprule
& Metric & \makecell{F-EX\\$\tau_c$} & \makecell{F-CF\\$\tau_b$} & \makecell{Com\\$\tau_c$} & \makecell{Pas\\acc.} \\
\midrule
\multirow{12}{*}{\rotatebox{90}{Reference-based}}
& BLEU & $30.8$ & $16.9$ & $30.6$ & $72.9$ \\
& ROUGE & $32.3$ & $19.9$ & $32.4$ & $74.1$ \\
& METEOR & $41.8$ & $22.2$ & $38.9$ & $78.0$ \\
& CIDEr & $43.9$ & $24.6$ & $37.7$ & $76.8$ \\
& SPICE & $44.9$ & $24.4$ & $40.3$ & $69.6$ \\
& BERT-S & $39.2$ & $22.8$ & $30.1$ & $79.1$ \\
& TIGEr & $49.3$ & -- & $45.4$ & $80.7$ \\
& ViLBERTS-F & $50.1$ & -- & $52.4$ & $79.6$ \\
& FAIEr-4 & $52.6$ & $35.4$ & $57.7$ & $81.4$ \\
& RefCLIP-Score & $53.0$ & $36.4$ & $55.4$ & $83.1$ \\
& Polos & $56.4$ & $37.8$ & $57.6$ & $\bm{86.5}$ \\
& RefFLEUR & $51.9$ & $38.8$ & $64.2$ & $85.5$ \\
\midrule
\multirow{10}{*}{\rotatebox{90}{Reference-free}}
& UMIC & $46.8$ & $30.1$ & $56.1$ & $85.1$ \\
& FAIEr-r & $50.1$ & $32.4$ & $50.5$ & -- \\
& CLIP-S & $51.5$ & $34.4$ & $53.8$ & $80.7$ \\
& InfoCLIP & $32.6$ & $23.5$ & $15.3$ & $64.1$ \\
& InfoMetIC & $54.2$ & $36.3$ & $59.2$ & $85.3$ \\
& InfoMetIC$^{+}$ & $55.5$ & $36.6$ & $59.3$ & $\bm{86.5}$ \\
& G-VEval & $\bm{59.7}$ & $38.7$ & $63.0$ & $82.3$ \\
& FLEUR & $53.0$ & $38.6$ & $63.5$ & $83.2$ \\
& GPT-FLEUR & $53.5$ & $39.0$ & $61.5$ & $82.6$ \\
& HarmonicEval & $53.1$ & $\bm{39.2}$ & $\bm{66.2}$ & $82.4$ \\ 
\bottomrule
\end{tabular}
\caption{Comparison on Flickr8k-EX / CF (F-EX/CF), Composite (Com), and Pascal-50S (Pas).}
\label{tab:score}
\end{table}

\begin{table}
\tabcolsep 5pt
\small
\centering
\begin{tabular}{lcc}
\toprule
Metric & 1-ref & 4-ref \\
\midrule
Polos & $93.3$ & $95.4$ \\
RefFLEUR & $97.3$ & $\bm{98.4}$ \\
\midrule
FLEUR & $96.8$ & $96.8$ \\ 
GPT-FLEUR & $97.0$ & $97.0$ \\ 
HarmonicEval & $\bm{97.8}$ & $97.8$ \\ 
\bottomrule
\end{tabular}
\caption{Comparison on FOIL.}
\label{tab:foil}
\end{table}

\section{Related Work}


\paragraph{Automatic evaluation metrics.}
Traditional metrics such as BLEU~\cite{Papineni2002bleu} and ROUGE~\cite{Lin2004rouge} were developed for automatic evaluation of natural language generation (NLG), relying on n-gram overlap with references.
Recently, large language models (LLMs) have been increasingly employed as evaluators across various NLG tasks~\cite{kocmi-federmann-2023-large, chiang-lee-2023-closer, NEURIPS2023_91f18a12, song-etal-2024-finesure}. For example, G-Eval~\cite{liu2023geval} introduced a form-filling paradigm for criterion-based evaluation in summarization and dialogue generation.
Our HarmonicEval extends this line by handling various multi-modal tasks and aggregating criterion-wise scores through second-order statistics of token probability distributions from VLMs.

For the IC task, CIDEr~\cite{Vedantam2015cider} measures the consensus between candidate and reference captions by weighting n-grams using TF-IDF.
Recent metrics leverage VLMs to offer more flexible evaluation paradigms~\cite{zhang2023gpt4visiongeneralistevaluatorvisionlanguage, lee-etal-2024-prometheus, yu2024mmvet, zhuang-etal-2024-automatic, maeda2024visionlanguagemodelbasedcaption, DBLP:conf/icml/ChenCZWLZZ00024, Tong_He_Shao_Yeung_2025}. FLEUR~\cite{lee-etal-2024-fleur}, a state-of-the-art reference-free metric, provides textual explanations that underlie its overall scores.
However, existing metrics primarily focus on providing the overall evaluation, and often overlook specific criteria. HarmonicEval addresses these issues by offering criterion-wise scores alongside the overall score, enabling comprehensive evaluation.

\paragraph{Human evaluation benchmarks.}
Several human evaluation benchmarks have been proposed for NLG tasks. Example benchmarks include SummEval~\cite{fabbri-etal-2021-summeval} for text summarization and the WMT shared tasks~\cite{semenov-etal-2023-findings} for machine translation.
In the multi-modal domain, existing benchmarks such as Flickr8k-EX / CF, Composite, PASCAL-50S, THUMB~\cite{kasai-etal-2022-transparent}, and Polaris~\cite{Wada2024PolosPolaris} target image captioning.
While some recent benchmarks~\cite{10.1007/978-3-031-72658-3_13, Li_2024_CVPR} cover multiple tasks, they mainly adopt multiple-choice settings that diverge from real-world text generation.
In contrast, our MMHE benchmark provides a multi-task, multi-criteria human evaluation resource, including sentences generated by several state-of-the-art VLMs across diverse tasks. MMHE enables more nuanced evaluations that capture a variety of criteria and tasks, advancing the field beyond the focus on captioning alone.

\section{Conclusion}
We introduced HarmonicEval, a novel reference-free evaluation metric for multiple multi-modal tasks. 
HarmonicEval predicts criterion-wise scores and aggregates them via a statistically principled method to produce an overall score. 
In addition, we constructed MMHE, the first multi-task multi-criteria human evaluation benchmark, consisting of 18,000 expert judgments.
Experimental results demonstrate that HarmonicEval aligns more closely with human judgments than existing metrics on both MMHE and other commonly used human evaluation benchmarks.
Furthermore, our analysis with MMHE reveals that existing metrics tend to prioritize certain criteria while neglecting others.

\section{Limitations}

This section discusses limitations from six perspectives:  theory, modality, criteria, data collection, model, and computational cost.

\paragraph{Evaluation Bias.} 
In HarmonicEval, since score distributions are approximated by the output token probabilities, there is a possibility of unintended bias.
Notably, evaluation bias in LLM-based evaluation metrics has been documented in several studies~\cite{NEURIPS2023_91f18a12, liu2023geval, ohi-etal-2024-likelihood}.
As such, further research into evaluation bias in VLM-based evaluation metrics is essential for future work.

\paragraph{Modality.} This study focused primarily on evaluating text quality in vision-language tasks because most state-of-the-art VLMs output only text. This leaves image quality evaluation underexplored. Given that several recent image generation models, such as DALL-E 3, are integrated into conversational systems using VLMs, the automatic evaluation of both generated image and text quality would be a promising next step toward the development of more user-friendly systems.

\paragraph{Criteria.}
We carefully selected five general criteria that are considered effective across various multi-modal tasks.
These criteria were useful for discussions spanning the four multi-modal tasks in this study.
To further expand research to include a greater number of criteria and tasks, analyzing the relationship between task- or domain-specific criteria and these general criteria would also be necessary in future work.

\paragraph{Data collection.}
As the number of criteria increased, it became difficult even for experts to maintain annotation consistency, leading to greater time requirements for data collection.
MMHE extracted one hundred images from each task, which was considered to be a statistically reliable number, and each target text was evaluated by three annotators.
While large-scale crowdsourcing was attempted to scale up this benchmark, obtaining human judgments that adhered accurately to each rating scale was challenging because careful explanation of the rating process through direct communication was required.
This leaves scaling up the number of tasks and images challenging.

\paragraph{Model.}
Improving VLMs to generate better text based on the evaluation results remains future work. In particular, achieving high scores across all criteria within learning frameworks such as in-context learning or reinforcement learning would be an intriguing direction for further exploration.

\paragraph{Computational cost.}
Since HarmonicEval relies on five prompts for criterion-wise scoring, it incurs five times the computational or API cost compared to simpler prompting methods.
We consider this a trade-off between achieving more robust and fine-grained metrics and managing computational cost.
This approach can also be seen as a form of \textit{test-time scaling}~\cite{snell2025scaling, xu2024llavacot}, where system performance is improved by increasing computational resources at inference time.
One potential direction to mitigate this limitation is to have the model generate scores for all criteria in a single output. However, we did not pursue this approach, as it complicates the computation of score expectations and variances. We leave this for future work.

\section{Ethics Statement}

\paragraph{Data collection.}
We created and will publicly release a new benchmark as part of this research.
The data collection process was conducted with careful consideration for ethical guidelines. All annotators were informed about the purpose of this benchmark and provided consent before participation.
Any personally identifiable information has been removed to ensure privacy protection.
The benchmark was reviewed to minimize harmful content, offensive language, or biases that could negatively impact downstream applications. However, some inherent biases might still be present due to the nature of the data sources including natural images.

\paragraph{Reproducibility.} 
All code necessary to reproduce the experimental results will be made publicly available. All experiments have been conducted as deterministically as possible by fixing random seeds and setting the temperature hyperparameters to zero.

\section{Acknowledgements}
This work was supported by JSPS KAKENHI Grant Number 25K03135. These research results were obtained from the commissioned research (No.22501) by National Institute of Information and Communications Technology (NICT) , Japan. This work was supported by the ``R\&D Hub Aimed at Ensuring Transparency and Reliability of Generative AI Models'' project of the Ministry of Education, Culture, Sports, Science and Technology. This study was carried out using the TSUBAME4.0 supercomputer at Institute of Science Tokyo.

\section{Bibliographical References}\label{sec:reference}

\bibliographystyle{lrec2026-natbib}
\bibliography{lrec2026-example}


\end{document}